\documentclass[letterpaper, 10 pt, conference]{ieeeconf}  

\IEEEoverridecommandlockouts                              

\usepackage{times}
\usepackage{epsfig}
\usepackage{graphicx}
\usepackage{amsmath}
\usepackage{amssymb}
\usepackage{gensymb}
\usepackage{subfig}
\usepackage{float}
\usepackage{mathtools}
\usepackage{dblfloatfix}
\usepackage{mathbbol}
\usepackage{subfig}
\usepackage{makecell}
\usepackage{romannum}
\usepackage[export]{adjustbox}
\usepackage{comment}
\usepackage{multirow}
\usepackage{array}
\usepackage{booktabs}
\usepackage{hyperref}

\usepackage[usernames,dvipsnames,svgnames,table]{xcolor}

\definecolor{fullred}{rgb}{0.85,.0,.1} 
\definecolor{cadmiumorange}{rgb}{0.93, 0.53, 0.18}
\definecolor{navyblue}{rgb}{.0,.0,.5}
\definecolor{bleudefrance}{rgb}{0.19, 0.55, 0.91}
\definecolor{bluegray}{rgb}{0.18, 0.36, 0.6}
\definecolor{lightgray}{rgb}{0.95, 0.95, 0.95}
\definecolor{darkgray}{rgb}{0.41, 0.41, 0.41} 
\definecolor{white}{rgb}{1.0, 1.0, 1.0}

\newcommand\darkgray[1]{\textcolor{darkgray}{#1}}

\newcommand{\rebuttal}[1]{\textcolor{black}{#1}}

\newcommand\ModelName{DD3Dv2}

\title{\LARGE \bf
Depth Is All You Need for Monocular 3D Detection
}

\author{Dennis Park$^*$, Jie Li$^*$, Dian Chen, Vitor Guizilini, Adrien Gaidon 
\thanks{*Equal Contribution}
\thanks{Toyota Research Institute,
         {\tt\small firstname.lastname@tri.global}}%
}

\begin{document}

\maketitle
\thispagestyle{empty}
\pagestyle{empty}

\begin{abstract}
A key contributor to recent progress in 3D detection from single images is monocular depth estimation.
\rebuttal{Existing methods focus on how to leverage depth explicitly by generating pseudo-pointclouds or providing attention cues for image features.}
More recent works leverage depth prediction as a pretraining task and fine-tune the depth representation while training it for 3D detection. However, the adaptation is insufficient and is limited in scale by manual labels.
In this work, we propose further aligning depth representation with the target domain in unsupervised fashions. Our methods leverage commonly available LiDAR or RGB videos during training time to fine-tune the depth representation, which leads to improved 3D detectors.
\rebuttal{Especially when using RGB videos, we show that our two-stage training by first generating pseudo-depth labels is critical because of the inconsistency in loss distribution between the two tasks.}
With either type of reference data, our multi-task learning approach improves over state of the art on both KITTI and NuScenes, while matching the test-time complexity of its single-task sub-network.


\end{abstract}

\section{INTRODUCTION}
Recognizing and localizing objects in 3D space is crucial for applications in robotics, autonomous driving, and augmented reality. Hence, in recent years monocular 3D detection has attracted substantial scientific interest \cite{park2021dd3d, manhardt2019roi, wang2021fcos3d, simonelli2019disentangling}, because of its wide impact and the ubiquity of cameras. However, as quantitatively shown in \cite{simonelli2020demystifying}, the biggest challenge in monocular 3D detection is the inherent ambiguity in depth caused by camera projection. Monocular depth estimation \cite{packnet, eigen2014depth, dorncvpr, godard2018digging2} directly addresses this limitation by learning statistical models between pixels and their corresponding depth values, given monocular images.

One of the long-standing questions in 3D detection is how to leverage advances in monocular depth estimation to improve image-based 3D detection. Pioneered by \cite{wang2019pseudo}, pseudo-LiDAR detectors \cite{you2019pseudo, ma2020rethinking, qian2020end} leverage monocular depth networks to generate intermediate pseudo point clouds, which are then fed to a point cloud-based 3D detection network. 
However, the performance of such methods is bounded by the quality of the pseudo point clouds, which deteriorates drastically when facing domain gaps. Alternatively, \cite{park2021dd3d} showed that by pre-training a network on a large-scale multi-modal dataset where point cloud data serves as supervision for depth, the simple end-to-end architecture is capable of \rebuttal{learning geometry-aware representation} and achieving state-of-the-art detection accuracy on the target datasets.

\begin{figure*}
\centering
\subfloat[\centering DD3Dv2: Depth Supervision Improves 3D Detection]{
\includegraphics[width=0.70 \textwidth]{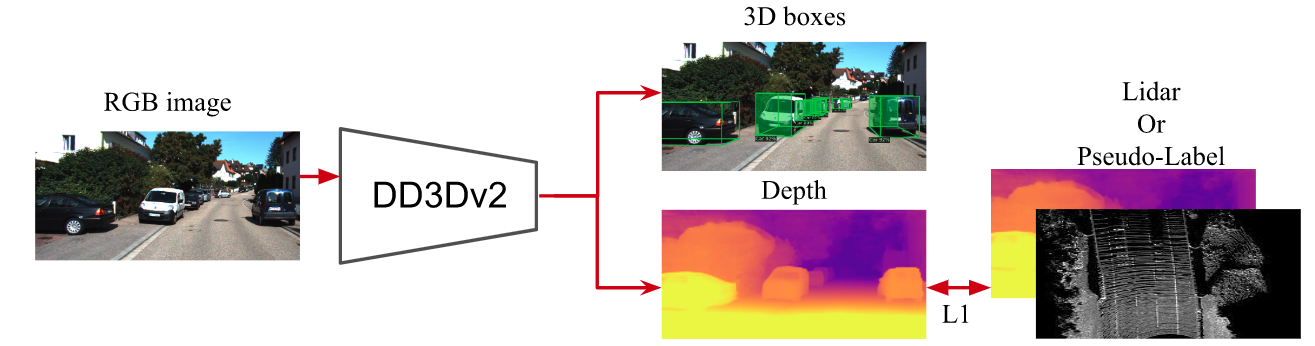} 
}
\subfloat[\centering Multi-task Head]{
\includegraphics[width=0.30 \textwidth]{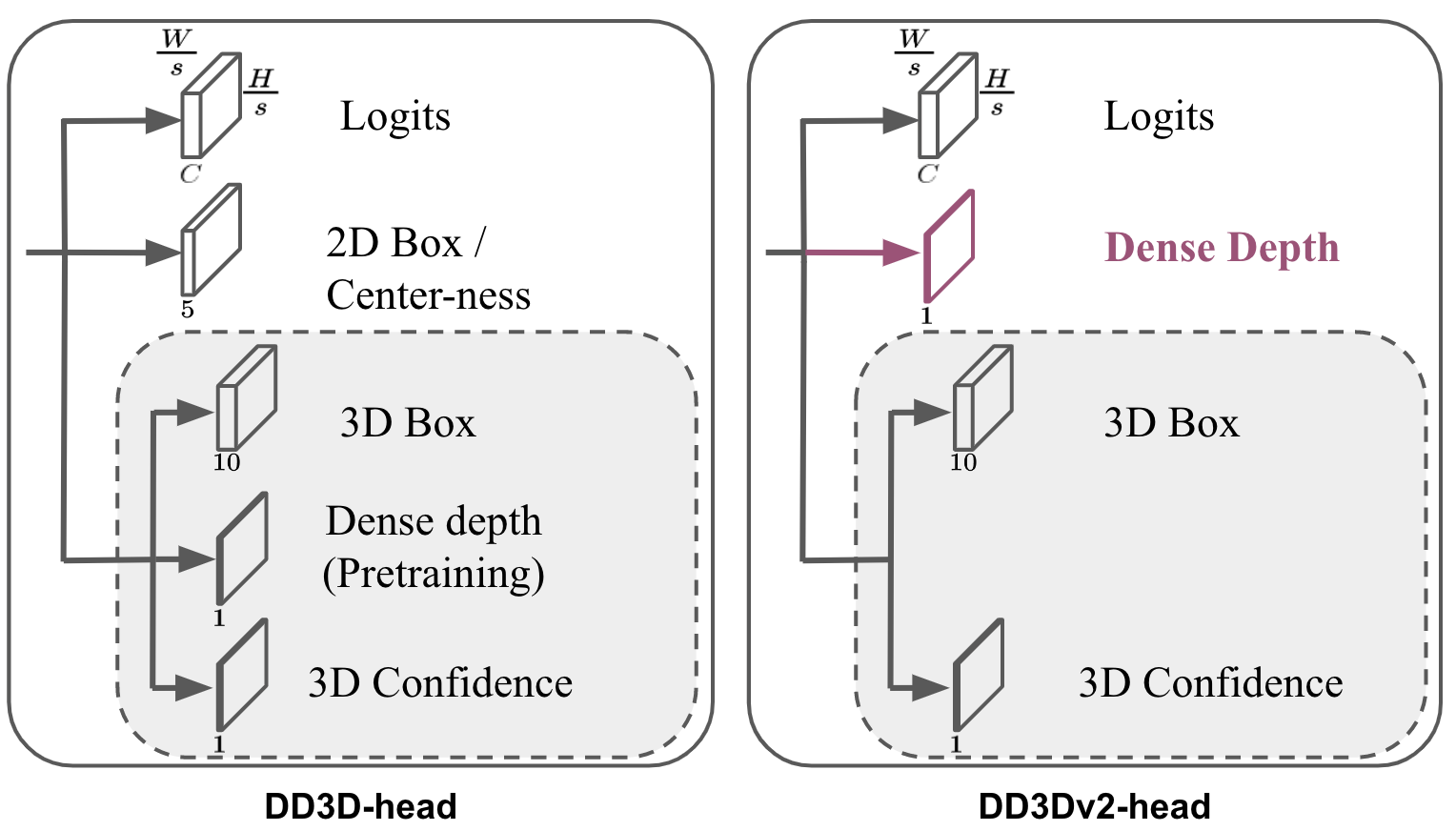} 
\label{fig:difference}
}
\caption{\rebuttal{DD3Dv2. This paper proposes a simple and effective algorithm to improve monocular 3D detection through depth supervision. (a): The overall flowchart of our proposed system can be adapted to both LiDAR supervision or Camera videos through pseudo labels generated from self-supervision algorithms. (b) Our multi-task decoder head improves on top of the original DD3D by removing redundant information streams.}}
\label{fig:teaser}
\vspace{-6mm}
\end{figure*}

However, in~\cite{park2021dd3d} the dataset used for pre-training exhibits a significant domain gap from the target data used for 3D detection. The source of this domain gap includes geographical locations (which affects scene density, weather, types of objects, etc) and sensor configuration (e.g. camera extrinsics and intrinsics). 
\rebuttal{It is unclear whether the geometry-aware representation learned during pretraining is sufficiently adapted to the new domain during fine-tuning.}
The goal of this work is to push the boundaries of how much pre-trained networks can be adapted for robust 3D detection using various types of unlabeled data available in the target domain.

We first consider scenarios where in-domain point cloud data is available at training time, sharing the assumptions with~\cite{dorncvpr, godard2018digging2}. In this case, we show that a simple multi-task framework supervised directly with projected depth maps along with 3D bounding boxes yields impressive improvements, compared with pseudo-LiDAR approaches \cite{you2019pseudo, ma2020rethinking} or pre-training based methods \cite{park2021dd3d}. Unlike pseudo-LiDAR methods, our methods entail no additional overhead at test time.

While it spawns insightful research ideas, the assumption that in-domain point cloud data is available during training can be impractical. For example, most outdoor datasets for 3D detection assume either multi-modal settings \cite{caesar2020nuscenes, Sun_2020_CVPR, geiger2012we} or a camera-only setting \cite{Neuhold_2017_ICCV, objectron2021} during both training and testing. Therefore, we propose an alternative variant to our method which adapts depth representations requiring only RGB videos. 

Inspired by advances in self-supervised monocular depth estimation \cite{packnet, eigen2014depth, zhou2018stereo}, we extend our method to using temporally adjacent video frames when LiDAR modality is not available. In this case, we observe that naively applying the same multi-task strategy with the two heterogeneous types of loss (2D photometric loss~\cite{eigen2014depth} and 3D box L1 distance),  results in sub-par performance. To address this heterogeneity, we propose a two-stage method: first, we train a self-supervised depth estimator using raw sequence data to generate dense depth predictions or \emph{pseudo-depth} labels. Afterward, we train a multi-task network supervised on these pseudo labels, using a distance-based loss akin to the one used to train the 3D detection. We show that this two-stage framework is crucial to effectively harness the learned self-supervised depth as a means for accurate 3D detection. 
In summary, our contributions are as follows:


\begin{itemize}
    \item \rebuttal{We propose a simple and effective multi-task network, DD3Dv2, to refine depth representation for more accurate 3D detection. Our method uses depth supervision from unlabelled data in the target domain during \emph{only} training time.} 
    \item \rebuttal{We propose methods for learning depth representation under two practical scenarios of data availability: LiDAR or RGB video. For the latter scenario, we propose a two-stage training strategy to resolve the heterogeneity among the multi-task losses imposed by image-based self-supervised depth estimation. We show that this is crucial for performance gain with empirical experiments.}
    \item \rebuttal{We evaluate our proposed algorithms in two challenging 3D detection benchmarks and achieve state-of-the-art performance.}
    

\end{itemize}

\section{RELATED WORK}\label{sec:related-work}

\subsection{Monocular 3D detection}
Early methods in monocular 3D detection focused on using geometry cues or pre-trained 3D representations to predict 3D attributes from 2D detections and \rebuttal{enforce 2D-3D consistency}~\cite{chabot2017manta,ansari2018graded,barabanau2019monocular,manhardt2019roi,ku2019monocular}. They often to require additional data to obtain geometry information, such as CAD models or instance segmentation masks at training time, and the resulting performance was quite limited.

Inspired by the success of point-cloud based detectors, a series of \emph{Pseudo-LiDAR} methods were proposed \cite{wang2019pseudo, weng2019monocular, qian2020end, wang2020train, ma2019accurate}, which first convert images into a point-cloud using depth estimators, and then apply ideas of point-cloud based detector. A clear advantage of such methods is that, in theory, a continuous improvement in depth estimation leads to more accurate detectors. However, the additional depth estimator incurs a large overhead in inference.


An alternative category is end-to-end 3D detection, in which 3D bounding boxes are directly regressed from CNN features~\cite{liu2020smoke,simonelli2019disentangling, wang2021fcos3d, park2021dd3d}. These methods directly regress 3D cuboid parameterizations from standard 2D detectors~\cite{tian2019fcos, ren2015faster}. While these methods tend to be simpler and more efficient, these methods do not address the biggest challenge of image-based detectors, the ambiguity in depth.  
DD3D~\cite{park2021dd3d} partially addresses this issue by pre-training the network on a large-scale image-LiDAR dataset.


Our work adopts the idea of end-to-end detectors, pushing the boundary of how far a good depth representation can help accurate 3D detection. Our key idea is to leverage raw data in the target domain, such as point clouds or video frames, to improve the learning of geometry-aware representation for accurate 3D detection.

\rebuttal{Other recent works trying to leverage dense depth or its uncertainty as explicit information for 3D lifting~\cite{CaDDN}, feature attention~\cite{zhang2022monodetr} or detection score~\cite{Lu_2021_ICCV}.}
MonoDTR~\cite{huang2022monodtr} shares a similar spirit with us in leveraging in-domain depth through multitask network. However, MonoDTR focuses on the use of the predicted depth to help query learning in a Transfomer-style detector~\cite{carion2020end}. 
\rebuttal{Compared to these methods, our method focuses on implicit learning of the depth information through proper supervision signal and training strategy. No additional module or test-time overhead is involved in the baseline 3D detector.}



\subsection{Monocular Depth Estimation}

Monocular depth estimation is the task of generating per-pixel depth from a single image. Such methods usually fall within two different categories, depending on how training is conducted. \emph{Supervised} methods rely on ground-truth depth maps, generated by projecting information from a range sensor (e.g., LiDAR) onto the image plane. The training objective aims to directly minimize the 3D prediction error. In contrast, \emph{self-supervised methods} minimize the 2D reprojection error between temporally adjacent frames, obtained by warping information from one onto another given predicted depth and camera transformation. A photometric object is used to minimize the error between original and warped frames, which enables the learning of depth estimation as a proxy task.  

Another aspect that differentiates these two approaches is the nature of learned features. Supervised methods optimize 3D quantities (i.e., the metric location of ground-truth and predicted point-clouds), whereas self-supervised methods operate in the 2D space, aiming to minimize reprojected RGB information.  Because of that, most \emph{semi-supervised} methods, that combine small-scale supervision with large-scale self-supervision, need ways to harmonize these two losses, to avoid task interference even though the task is the same. In \cite{packnet-semisup}, the supervised loss is projected onto the image plane in the form of a reprojected distance, leading to improved results relative to the naive combination of both losses. In this work, we take the opposite approach and propose to revert the 2D self-supervised loss back onto the 3D space, through pseudo-label.

\section{MULTI-TASK LEARNING FOR 3D DETECTION}\label{sec:dd3dv2}
In this section, we introduce our multitask framework to adapt geometry-ware features in the target domain during training. While our proposed approach can be generalized to any end-to-end 3D detector (E.g. ~]\cite{liu2020smoke,wang2021fcos3d}), we build our model on top of DD3D~\cite{park2021dd3d} as a baseline. 
We briefly recapitulate DD3D and highlight our modifications to facilitate in-domain depth feature learning in our model, \emph{\ModelName}, as also depicted in Figure~\ref{fig:difference}. 

\textbf{DD3D Baseline} DD3D \cite{park2021dd3d} is a fully convolutional network designed for 3D detection and pre-training supervised by point-cloud data. The backbone network transforms the input image to a set of CNN features with various resolutions. The CNN features are then processed by three different \emph{heads}, each comprising $4$ of $3\times3$ convolutional layers and compute logits and parameterizations of 2D  / 3D boxes. We refer the readers to \cite{park2021dd3d} for more detail on the architecture and decoding schemes.

\textbf{Depth head.} The design of a shared head for depth and 3D box prediction in DD3D is motivated by enhancing knowledge transfer between the (depth) pre-training and detection. However, in the scenario of multi-task, we found that excessive sharing of parameters causes unstable training. Therefore, we keep the parameters for depth prediction as an independent head with the same architecture of other heads, which consists of $4$ of $3\times3$ convolution layers. 

\textbf{Removal of 2D box head.} Adding an additional head incurs significant overhead in memory and hinders large-scale training with high-resolution images. Since we are only interested in 3D detection, we remove the 2D box head and center-ness. The 2D boxes used in non-maxima suppression are replaced by axis-aligned  boxes that tightly contain the projected key points of 3D boxes. This results in a three-head network, with similar memory footprints of DD3D.


\textbf{Improved instance-feature assignment.} When training fully convolutional detectors, one must decide how to associate the ground-truth instances to the predicted candidates. DD3D adopts a CenterNet-style \cite{zhou2019objects} strategy that matches the centers of ground-truth 2D boxes with the feature locations. However, applying this method to multi-resolution features (e.g. FPN \cite{lin2016fpn}) causes a boundary effect between scale ranges. Instead of applying hard boundaries in scale space, we adopt a strategy of using \emph{anchor boxes} (i.e. 2D boxes with various sizes and aspect ratios centered at a feature location) associated with features to determine the assignments. Given a feature location $l$ and a ground-truth bounding box $\mathcal{B}_{g}=(x_1, y_1, x_2, y_2)$, the matching function $\mathcal{M}$ is defined as:
\begin{align}
    \mathcal{M}(l, \mathcal{B}_g) = I[\max_{\mathcal{B}_a \in \mathcal{A}(l)} v(\mathcal{B}_a, \mathcal{B}_g) > \tau]
\end{align}
where $\mathcal{A}(l)$ is a set of anchor boxes associated with the location $l$, $v(\cdot, \cdot)$ is an overlapping criteria (e.g. IoU), and $\tau$ is a threshold. 

This effectively produces a \emph{soft} boundary between the scale ranges and allows for many-to-one assignments. We observed that this leads to more stable training. On nuScenes validation split, this modification leads to a significant improvement in detection accuracy, from 38.9\% to 41.2\% mAP.
\section{LEARNING DEPTH REPRESENTATION}
\label{sec:multitask}
\rebuttal{In this section, we describe how \emph{\ModelName} can be trained under different in-domain data availability.}

\subsection{Using point cloud}~\label{sec:lidar-supervision}
When point cloud data is available, we directly use it as supervision for the depth head in our multi-task training.  Following \cite{park2021dd3d}, we first project the point cloud onto the image plane and calculate smoothed L1 distance on the pixels with valid ground truth. Camera intrinsics are used to re-scale the depth prediction to account for variable input resolutions caused by data augmentation \cite{park2021dd3d}. 

\subsection{Using camera video}
\begin{figure*}[h!]
    \centering
    \subfloat[\centering Single stage strategy]{{\includegraphics[width=0.4 \textwidth]{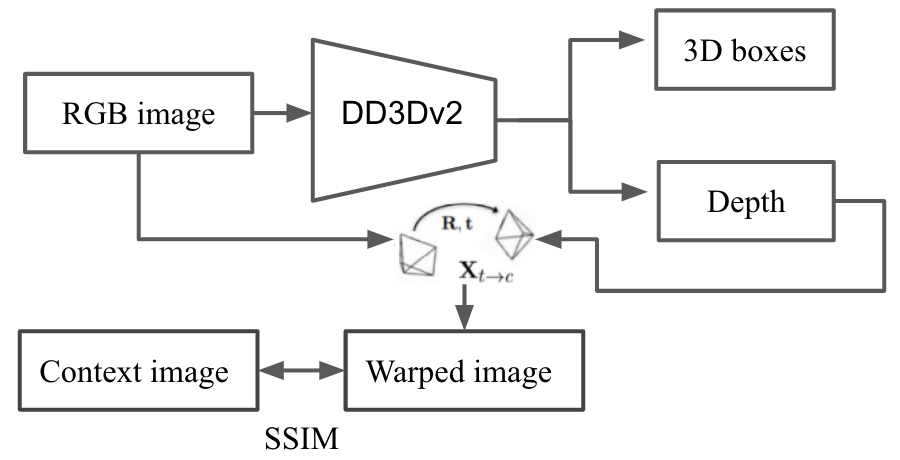} }}%
    \hspace{3pt}
\subfloat[\centering Two stages strategy (Proposed)]{{\includegraphics[width=0.52 \textwidth]{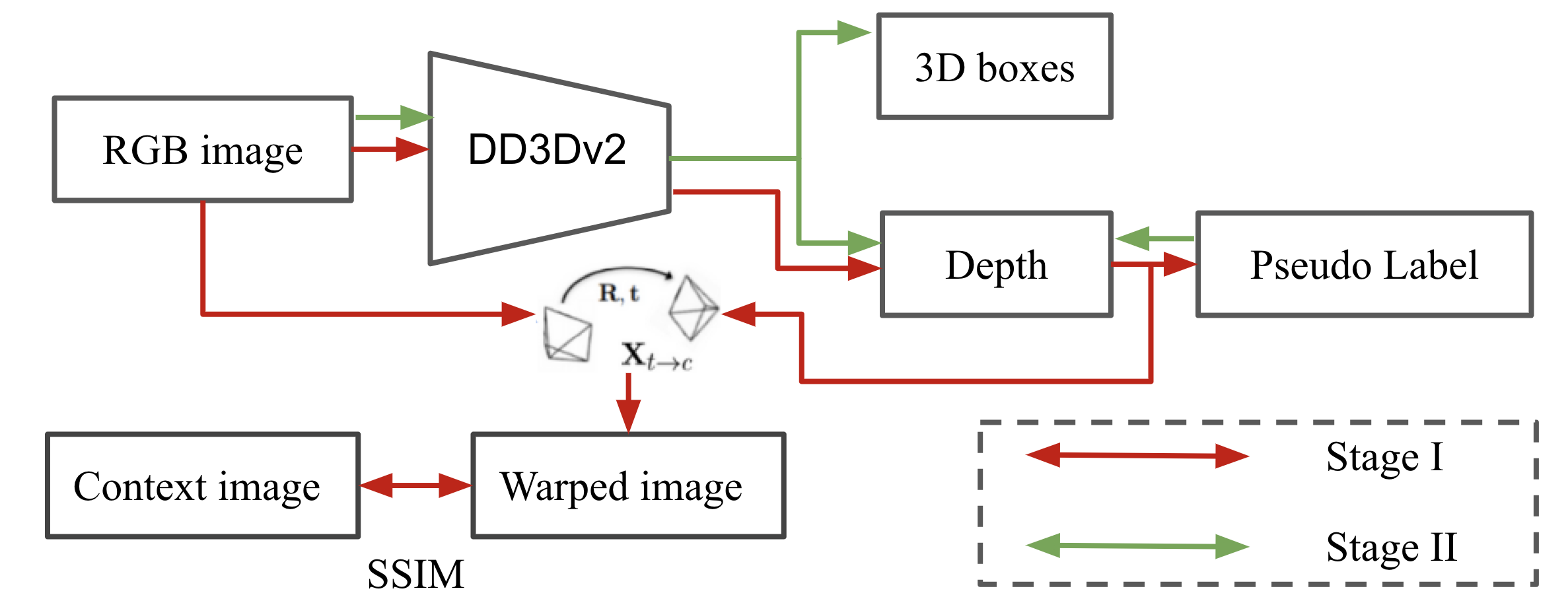} }}%
    \caption{\rebuttal{To use self-supervision techniques to guide depth supervision using context images from the video, we discuss two training strategies here. (a) The most straight forward and convenient strategy would be directly combine the self-supervised training paradigm as part of the multi-task network. (b) The second strategy would be first train a depth network that can be used to generate pseudo ground truth depth. Then apply multi-task training in the second stage using pseudo label the same way we use LiDAR. We found that the second strategy provide more significant improvement to the original 3D detection compared to the first one.}}
    \label{fig:training-strategy}
\end{figure*}
Given video frames instead of point cloud data, we adopt a two-stage pseudo-label framework. Concretely, \rebuttal{as depicted in Figure~\ref{fig:training-strategy}(b)}, we first learn a depth network on the target data via self-supervised depth estimation (\cite{packnet, eigen2014depth}) in stage I, and then train our multi-task network using pseudo depth labels generated from the learned depth network. Stage II is similar to Sec.~\ref{sec:lidar-supervision}, but the target (pseudo) depth labels are dense compared to LiDAR point clouds.


\rebuttal{\textbf{Single stage vs. Two-Stage}}
\rebuttal{Given video frames, the most direct and computationally efficient way to use it with DD3Dv2 is to adopt the same multi-task training, substituting the direct depth supervision with self-supervised photometric loss ~\cite{zhou2017unsupervised} (Fig. ~\ref{fig:training-strategy}(a)). We refer to it as the single-stage strategy for the rest of the paper.  }

\rebuttal{The photometric loss substitutes the direct depth estimation error with reprojection error in RGB space between two images: the target image on which the pixel-wise depth is estimated $I_t$,  and the synthesized version of it formed by warping the neighboring frames $\hat{I}_t$}. The difference in appearance is measured by SSIM \cite{ssim} and L1 distance of (normalized) pixel values:
\begin{equation}
    \mathcal{L}_p(I_t, \hat{I}_t) = \alpha \frac{1 - \text{SSIM}(I_t, \hat{I_t})}{2} + (1 - \alpha) || I_t - \hat{I_t} || 
\end{equation}

\rebuttal{While photometric loss has been widely adopted in most of the self-supervised monocular depth estimation works~\cite{zhou2017unsupervised, packnet}, we found that it does not work compatibly with direct 3D losses used in 3D detection, as demonstrated in Table~\ref{tab:ablation} (E3, E4 vs E1). }

\rebuttal{For 3D detection optimization, we apply disentangling 3D boxes loss computation~\cite{simonelli2020disentangling} on 3D box loss to optimize 3D box components independently (orientation, projected center, depth, and size).
\begin{equation}
    \mathcal{L}_{\text{3D}}(\mathbf{B}^*, \mathbf{\hat{B}}) =  ||\mathbf{B}^* - \hat{\mathbf{B}}||_1 ,
\end{equation}
where ground truth for other components is provided when the targeted component is being evaluated.
In the case of depth, the 3D box loss equals a simple L1 loss.}

\begin{figure*}
\centering
\subfloat[\centering RGB image]{{\includegraphics[width=0.30 \textwidth]{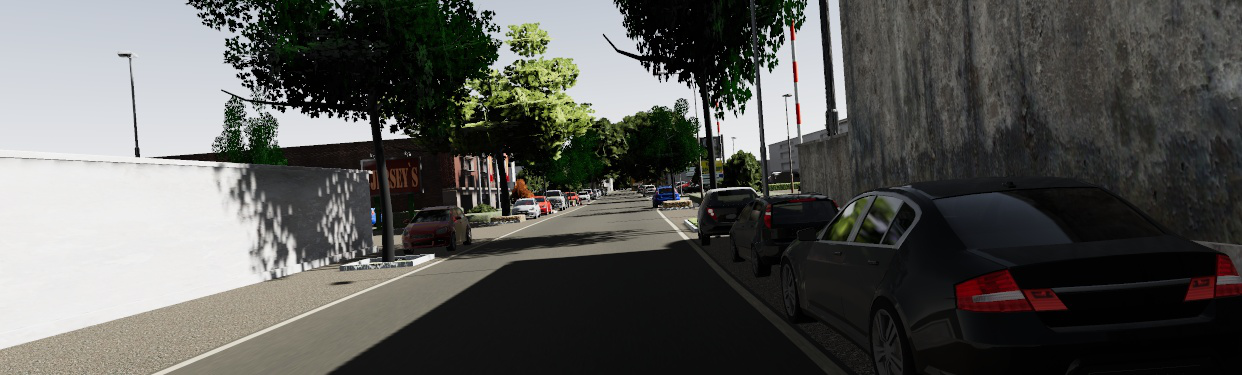} }}%
\subfloat[\centering L1 Loss on depth ]{{\includegraphics[width=0.30 \textwidth]{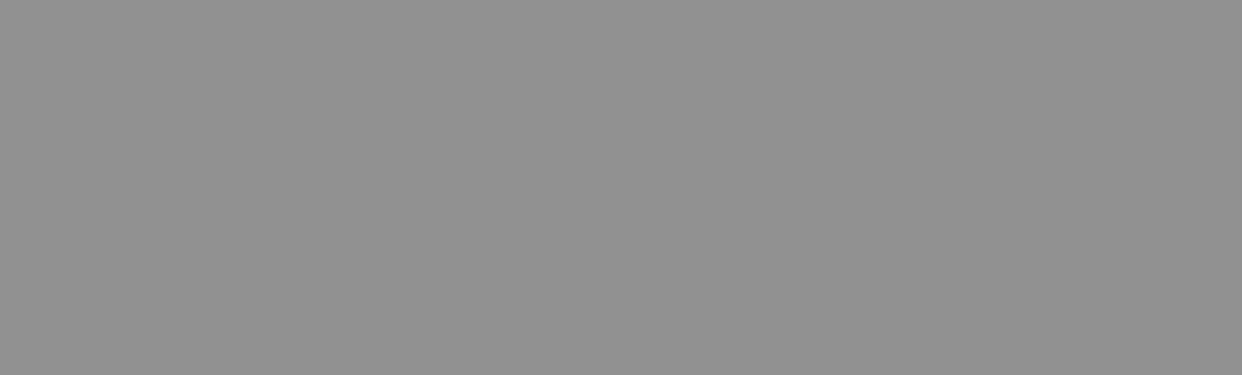} }}%
\subfloat[\centering Photometric loss]{{\includegraphics[width=0.30 \textwidth]{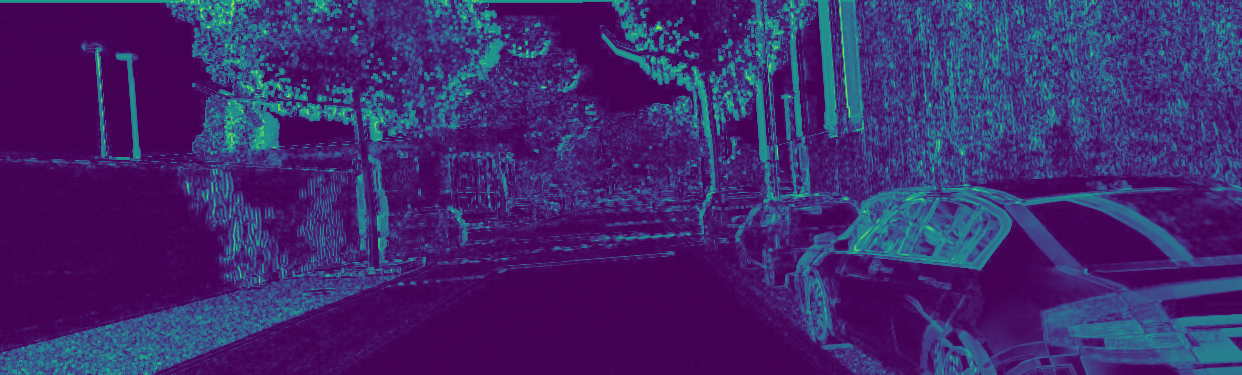} }}%

\caption{\rebuttal
{Heterogeneity between photometric loss and L1 loss. We provide an illustration of the two loss distribution when depth prediction error is $\textbf{1m}$ everywhere. While $L1$ loss is a direct measurement of depth estimation, photometric loss is also correlate with the structure and appearance in the scene. (This figure use perfect ground truth from vKITTI~\cite{gaidon2016virtual} dataset.)}}  

\label{fig:selfsup_loss}
\end{figure*}

\rebuttal{In the single-stage strategy, this heterogeneity of the two losses causes a large difference in the distribution of depth prediction and its error. In Figure~\ref{fig:selfsup_loss}, we visualize these losses to better illustrate this heterogeneity. }

Compare to L1 loss, the photometric loss is correlated with the structure and the appearance of the scene. It exhibits different patterns depending on the distance of the object or structure in a scene. For example, objects further away or towards the vanishing point will be less sensitive to the depth error, due to a decrease in pixel resolution. A similar observation is also discussed in ~\cite{packnet-semisup}.

To address this inconsistency, we propose to use the self-supervised depth network in a similar fashion to how we use point-cloud data. Namely, we apply the self-supervised network to training data to obtain \emph{pseudo} depth labels, which are used in the same way as LiDAR point cloud to train the multi-task network with L1 loss. In this way, the depth loss shares the L1 nature (i.e. distance in 3D scenes) as detection loss. This yields improvement in 3D detection (Sec. \ref{sec:ablation}).


\section{BENCHMARK RESULTS}
\subsection{Datasets}
\noindent\textbf{nuScenes.} The nuScenes dataset \cite{caesar2020nuscenes} contains $1000$ videos divided into training, validation and test splits with $700$, $150$, and $150$ \emph{scenes}, respectively. Each sample is composed of $6$ cameras covering the full $360$-degree field of view around the vehicle, with corresponding annotations. 
The evaluation metric, \emph{nuScenes detection score} (\emph{NDS}), is computed as a linear combination of mean average precision \emph{mAP} over four thresholds on center distance and five \emph{true-positive} metrics. We report NDS and mAP, along with the three true-positive metrics that concern 3D detection, i.e. \emph{ATE}, \emph{ASE}, and \emph{AOE}. 

\noindent\textbf{KITTI-3D.} The KITTI-3D benchmark \cite{geiger2012we} contains a training set of $7481$ images and a test set of $7518$ images. 
For the 3D detection task, three object classes are evaluated on two average precision (AP) metrics: \emph{3D AP} and \emph{BEV AP}, which use intersection-over-union criteria on (projected) 3D boxes. The metrics are computed on three difficulty levels:  Easy, Moderate, and Hard.



\subsection{Implementation Details}
In all experiments, we initiate our model using pretrained weights (V2-99) from \cite{park2021dd3d}. We use the SGD optimizer with a learning rate of $2\times10^{-3}$, momentum of $0.9$ and weight decay at $1\times10^{-4}$, and batch size of $64$. For nuScenes, we train our model for $120K$ iterations with multi-step scheduler that decreases the learning rate by $10$ at steps $100K$ and $119K$. For KITTI, we train for $35K$ iterations and similarly decrease the learning rate at $32K$ and $34K$ steps. Ground truth poses are used in self-supervised depth training.
\subsection{Discussion}

In Table~\ref{table:nuscenes_test_summary}, we compare our model with published monocular approaches. (We exclude entries that use temporal cues at test time.) DD3Dv2, when trained using point-cloud supervision, yields higher accuracy than all other methods, including recent Transformer-based methods. When trained using video frames, it performs competitively with other methods, and shows impressive improvement over DD3D.





\begin{table*}[h]
\centering
{
\tiny
\footnotesize
\setlength{\tabcolsep}{0.3em}
\rowcolors{2}{lightgray}{white}
\begin{tabular}{l|c|c|c|c|c|c|c}
\toprule
Methods & Depth Sup. &Backbone& AP[\%]$\uparrow$ & ATE[m]$\downarrow$  & ASE[1-IoU]$\downarrow$ & AOE[rad]$\downarrow$ &  NDS$\uparrow$  \\
\midrule



MonoDIS~\cite{simonelli2020disentangling}  & - & R34&
30.4 & 
0.74 & 
0.26 & 
0.55 & 
0.38 \\

FCOS3D~\cite{wang2021fcos3d}            &  -  &R101 &
{35.8} & 
0.69 & 
{0.25} & 
{0.45} & 
0.43 \\
PGD\cite{wang2022probabilistic}                           &- & R101 &
37.0 & 
0.66 & 
0.25 & 
0.49 & 
0.43 \\
DD3D ~\cite{park2021dd3d}               &  - & V2-99&
{41.8} & 
0.57 & 
{0.25} &
0.37 & 
0.48 \\

DETR3D~\cite{detr3d}                                 
&  - &V2-99 &
{41.2} & 
{0.64} & 
{0.26} &
{ 0.39} & 
0.48 \\
\midrule
BEVDet$^*$~\cite{huang2021bevdet}                                 
&  - & V2-99&
{42.4} & 
\textbf{0.52} & 
\textbf{0.24} &
0.37 & 
{0.49} \\
BEVFormer-S$^{*}$~\cite{li2022bevformer}                                 
&  - &V2-99 &
{43.5} & 
{0.59} & 
{0.25} &
{ 0.40} & 
0.50 \\
PETR$^*$~\cite{liu2022petr}                                 
&  - &V2-99 &
44.1 & 
{0.59} & 
{0.25} &
{ 0.38} & 
0.50 \\
\midrule
DD3Dv2-selfsup 
& Video & V2-99 &
{43.1} &
{0.57} &
{0.25} &
0.38 &
0.48 \\
DD3Dv2 
&LiDAR   & V2-99 & 
\underline{\textbf{46.1}}  &  \underline{\textbf{0.52}} &  \underline{\textbf{0.24}} & \underline{\textbf{0.36}} & \underline{\textbf{0.51}} \\



 
 \bottomrule
\end{tabular}\\\vspace{0.5mm}
\caption{
\textbf{nuScenes detection \emph{test} set evaluation.} We present summary metrics of the benchmark. * denotes results reported on the benchmark that do not have associated publications at the time of writing. The \textbf{bold} and \underline{underline} denote the best of all and the best excluding concurrent work, respectively. Note that PointPillars~\cite{lang2019pointpillars} is a Lidar-based detector.
}
\label{table:nuscenes_test_summary}
}
\end{table*}

\begin{table*}[t!]
\centering
{
\footnotesize
\rowcolors{2}{lightgray}{white}
\begin{tabular}{l|
c|ccc|ccc}
\toprule
& & \multicolumn{6}{c}{Car} \\
& &\multicolumn{3}{c}{BEV AP} & \multicolumn{3}{c}{3D AP} \\ 
 \multirow{-2}{*}{Methods}& \multirow{-2}{*}{Depth  Sup.}&
Easy & 
Med &
Hard &
Easy & Med & Hard \vspace{0.5mm}\\
\midrule






SMOKE~\cite{liu2020smoke} & - &
20.83 & 
14.49 &
12.75 &
14.03 &
9.76 &
7.84
\\

MonoPair~\cite{chen2020monopair} & -&
19.28 & 
14.83 &
12.89 &
13.04 &
9.99 &
8.65
\\

AM3D~\cite{ma2019accurate} & LiDAR&
25.03 & 
17.32 &
14.91 &
16.50 &
10.74 &
9.52
\\

PatchNet$\dagger$ ~\cite{ma2020rethinking} & LiDAR&
22.97 & 
16.86 &
14.97 &
15.68 &
11.12 &
10.17
\\

RefinedMPL~\cite{vianney2019refinedmpl} & &
28.08 & 
17.60 &
13.95 &
18.09 &
11.14 &
8.96
\\

D4LCN~\cite{ding2020learning} & LiDAR&
22.51 & 
16.02 &
12.55 &
16.65 &
11.72 &
9.51
\\

Kinematic3D~\cite{brazil2020kinematic} & Video &
26.99 & 
17.52 &
13.10 &
19.07 &
12.72 &
9.17
\\


Demystifying~\cite{simonelli2020demystifying} &  LiDAR&
- & 
- &
- &
{23.66} &
{13.25} &
{11.23}
\\

CaDDN~\cite{CaDDN} & LiDAR &
27.94 & 
18.91 &
17.19&
19.17 &
13.41 &
11.46
\\
MonoEF~\cite{zhou2021monocular} &Video &
29.03 & 
19.70 &
17.26&
21.29 &
13.87 &
11.71
\\

MonoFlex~\cite{zhang2021objects} & - &
28.23 & 
19.75 &
16.89&
19.94 &
13.89 &
12.07
\\
GUPNet~\cite{lu2021geometry} & - &
- & 
- &
-&
20.11 &
14.20 &
11.77

\\
PGD~\cite{wang2022probabilistic} & - &
30.56 & 
23.67 &
20.84&
24.35&
18.34 &
16.90
\\
DD3D~\cite{park2021dd3d} & - &
{30.98} & 
{22.56} &
{20.03} &
{23.22} &
{16.34} &
{14.20}
\\
\midrule
MonoDTR$^{\star}$~\cite{huang2022monodtr} & LiDAR &
28.59 & 
20.38 &
17.14&
21.99 &
15.39 &
12.73
\\
PS-fld$^{\star\dagger}$~\cite{Chen_2022_CVPR} & LiDAR &
{32.64} & 
{23.76} &
{20.64} &
{23.74} &
\textbf{{17.74}} &
{15.14}
\\
MonoDDE$^\star$~\cite{li2022diversity} & - &
{33.58} & 
{23.46} &
{20.37} &
{23.74} &
{17.14} &
{15.10}
\\
\midrule
Ours & LiDAR&
\underline{\textbf{35.70}} & 
\underline{\textbf{24.67}} &
\underline{\textbf{21.73}} &
\underline{\textbf{26.36}} &
\underline{17.61}&
\underline{\textbf{15.32}}
\\


\bottomrule
\end{tabular}\\
\vspace{0.5mm}
\caption{
\textbf{KITTI-3D \textit{test} set evaluation on \emph{Car}.} We report AP$|_{R_{40}}$ metrics. 
$^\star$ indicates concurrent works. $^\dagger$ indicates the usage of the \textbf{KITTI-depth} dataset, with a known information leakage between training and validation splits \cite{simonelli2020demystifying}. \textbf{Bold} and \underline{underline} denote the best of all and the best excluding concurrent work.
}
\label{table:kitti_3d_test}
}
\end{table*}

In Table~\ref{table:kitti_3d_test} and \ref{table:kitti_3d_test_multi_class}, we show the results on KITTI-3D benchmark. We report our results with point-cloud supervision, since KITTI allows for only a single submission. (Comparison of self-supervised depth is provided in supplemental material.) 
DD3Dv2 achieves the state-of-the-art in most metrics across all three categories when compared with most published and concurrent works, including the ones that uses similar point-cloud supervision and Pseudo-LiDAR approaches.
%
Our new representation significantly improves over end-to-end approaches like~\cite{park2021dd3d}, especially on smaller objects.


\begin{table*}[t]
\centering
{
\footnotesize
\setlength{\tabcolsep}{3.5pt}
\rowcolors{2}{lightgray}{white}
\begin{tabular}{l|ccc|ccc|ccc|ccc}
\toprule
& \multicolumn{6}{c}{Pedestrian} & \multicolumn{6}{c}{Cyclist} \\
& \multicolumn{3}{c}{BEV AP} & \multicolumn{3}{c}{3D AP} & \multicolumn{3}{c}{BEV AP} & \multicolumn{3}{c}{3D AP} \\ 
\multirow{-3}{*}{Methods}& 
Easy & 
Med &
Hard &
Easy & Med & Hard & Easy & 
Med &
Hard &
Easy & Med & Hard\\
\midrule


M3D-RPN~\cite{brazil2019m3d} &
5.65 &
4.05 &
3.29 &
4.92 &
3.48 &
2.94 &
1.25 &
0.81 &
0.78 &
0.94 &
0.65 &
0.47 \\

MonoPSR~\cite{ku2019monocular} &
7.24 &
4.56 &
4.11 &
6.12 &
4.00 &
3.30 &
{9.87} &
{5.78} &
{4.57} &
{8.37} &
{4.74} &
{3.68} \\


CaDDN~\cite{CaDDN}&
{14.72} &
{9.41} &
{8.17} &
{12.87} &
{8.14} &
{6.76} &
9.67 &
5.38 &
4.75 &
7.00 &
3.41 &
3.30 \\

DD3D &
{15.90} &
{10.85} &
{8.05} &
{13.91} &
{9.30} &
{8.05} &
3.20 &
{1.99} &
{1.79} &
{2.39} &
{1.52} &
{1.31} \\
\midrule
MonoDTR$^{\star}$~\cite{huang2022monodtr}&
{16.66} &
{10.59} &
{9.00} &
{15.33} &
{10.18} &
{8.61} &
5.84 &
{4.11} &
{3.48} &
{5.05} &
{3.27} &
{3.19} \\
MonoDDE$^*$~\cite{li2022diversity}&
{12.38} &
{8.41} &
{7.16} &
{11.13} &
{7.32} &
{6.67} &
6.68 &
{4.36} &
{3.76} &
{5.94} &
{3.78} &
{3.33} \\
PS-fld$^{\star\dagger}$~\cite{Chen_2022_CVPR}&
{\textbf{19.03}} &
\textbf{{12.23}} &
\textbf{{10.53} }&
\textbf{{16.95}} &
\textbf{{10.82}} &
\textbf{{9.26}} &
\textbf{12.80} &
\textbf{7.29} &
\textbf{6.05} &
\textbf{11.22} &
\textbf{6.18} &
\textbf{5.21} \\
\midrule
Ours &
\underline{17.74} &
\underline{12.16} &
\underline{10.49} &
\underline{16.25} &
\underline{10.82} &
\underline{9.24} &
     \underline    {10.67} &
\underline{7.02} &
\underline{5.78} &
\underline{8.79} &
\underline{5.68} &
\underline{4.75} \\
\bottomrule
\end{tabular}\\
\caption{
\textbf{KITTI-3D \textit{test} set evaluation on \emph{Pedestrian} and \emph{Cyclist}}. $^\star$ indicates concurrent works.  $^\dagger$ indicates the usage of the \textbf{KITTI-depth} dataset. \textbf{Bold} and \underline{underline} denote the best of all and the best excluding concurrent work.
}
\label{table:kitti_3d_test_multi_class}
}
\end{table*}


\section{ABLATION ANALYSIS}
\label{sec:ablation}

\begin{table*}[h]
\centering
{
\begin{tabular}{c|l|c|c|c|c
|c}
ID&\thead{Approach} & \thead{Extra Data} &  \thead{Pseudo\\Labels}  &\thead{Depth\\Loss} & \thead{Detection Accuracy \\ NDS $\uparrow$ \darkgray{(mAP [$\%$]$\uparrow$)}} & \thead{Depth Accuracy \\ Abs. Rel $\downarrow$} \\
\midrule
E1&\small{Detection Only} & - & - &  L1 & 41.2 \darkgray{(35.8)} & - \\ 
\midrule
\midrule
E2&DD3Dv2& LiDAR & -& L1 & \textbf{45.6} \darkgray{(39.1)} & 0.20 \\ 
\midrule
E3&Self-supervised & Video &-& SSIM  & 42.8 \darkgray{(36.4)} & 0.51 \\
%
E4&\quad \small{+ ignore close} &Video& - & SSIM & 42.9 \darkgray{(37.5)}  & 0.54 \\

\midrule
%
%
E5&DD3Dv2-selfsup & Video & $\surd$ & L1  & 43.2 \darkgray{(37.7)} & 0.51 $\rightarrow$ 0.52 \\
%
E6&\quad \small{+ ignore close} & Video & $\surd$ & L1  & \underline{43.7} \darkgray{(36.9)} &0.54 $\rightarrow$  0.54 \\
\end{tabular}

\vspace{1em}

\caption{\rebuttal{We provide an ablation analysis on crucial design choices of both architecture and training strategies. We show how LiDAR supervision improves on top of single-task training (E2 vs. E1). In E3 and E4, we employ a single-stage training strategy using video frames as depicted in Figure~\ref{fig:training-strategy}(a). In E5 and E6, we employ a two-stage training strategy by generating pseudo-labels first as depicted in Figure~\ref{fig:training-strategy}(b). ``ignore close" indicate a small trick to ignore closest depth estimation in self-supervised training. All methods start from a single initial model pretrained by large-scale depth supervision available from \cite{park2021dd3d}.}}

\label{tab:ablation}
}
\vspace{-5mm}
\end{table*}

\noindent\textbf{Experimental setup.} For ablative study, we use nuScenes dataset (\emph{train} and \emph{validation}). 
\rebuttal{To cover a wide range of variations, we adopt a lightweight version of the full training protocol with half training steps and batch size.}
The reduced training schedule causes degradation in detection accuracy of baseline detection-only DD3Dv2 model from $41.1\%$ to $35.8\%$ mAP. To understand the interplay between detection and depth accuracy, we also report depth metrics computed \emph{only} on foreground regions. 
\paragraph{Is supervised depth using point cloud data effective?}
With direct supervision for depth estimation task,  \rebuttal{E2} achieves clear improvement compared to \rebuttal{E1}. This supports our argument that 
\rebuttal{even without a significant change of architecture or explicit use of depth prediction,} the representation for 3D detection can be  significantly improved by adapting to a good depth representation.
over E5 shows that our multi-task training successfully acts as a means of domain adaptation.
\begin{table*}[h]
\centering
{
\begin{tabular}{l|c|c|c|c|c}
\thead{Backbone} & \thead{Multi-task}&\thead{Pretrained Dataset} & \thead{Pretrained Task}&  \thead{NDS $\uparrow$ } & \thead{mAP [$\%$]$\uparrow$}\\
\midrule
V2-99 & - & DDAD15M&Depth Est.& 41.2  & 35.8 \\ 

V2-99& $\surd$ & DDAD15M & Depth Est.&{45.6} (+\textbf{4.4})  & 39.1 (+\textbf{3.3})\\ 
\midrule
V2-99& - & COCO & 2D Det.& 40.8 & 34.0 \\ 
V2-99& $\surd$ & COCO & 2D Det.& 43.1 (+2.3)& 36.2(+2.2)\\ 
\midrule
\end{tabular}
\vspace{1em}
\caption{\rebuttal{We analyzed the relationship between the pretraining backbone and proposed the in-domain multi-task representation learning using depth supervision. We compare the same backbone training on COCO~\cite{lin2014microsoft} on 2D detection. (Released by ~\cite{lee2019centermask}.) The multi-task training paradigm is consistently improving over the detection-only case. It is also noticeable that geometry-aware backbones (pretrained on depth estimation) achieve more significant improvement than object-aware backbones (COCO).}} 

\label{tab:gen}
}
\vspace{-5mm}
\end{table*}
\paragraph{Are pseudo-labels necessary for self-supervised depth?}
When the supervision of depth is replaced by the self-supervision from video frames, we observe a clear loss in accuracy (E3/E4 compared to E1), and it only yields a mediocre improvement over the DD3Dv2 single task baseline. This gap is noticeably closed by training on the pseudo-labels (E5 vs. E3, E6 vs. E4). The pseudo-labels significantly reduce the gap from the naive multi-task training. We argue that removing the heterogeneity in the combined loss results in a better adaptation.

\paragraph{When does depth supervised Multi-task work?}
\rebuttal{
To better understand and evaluate the generalizability of the proposed training paradigm, we analyze the effectiveness of Lidar supervision against different pretraining conditions in Table~\ref{tab:gen}. We compare the geometry-aware backbone (DD3D15M~\cite{park2021dd3d}) and objectness-aware backbone (COCO~\cite{lin2014microsoft} released by~\cite{lee2019centermask}). From both of the pretraining weights, multi-task learning with dense depth supervision can improve 3D detection by a clear margin. The geometry-aware model sees a higher improvement ($4.4$ over $2.3$ NDS), which further verifies our intuition that the multi-task training improves the adaptation of the geometry information in the pretrained weights into the target domain. 
}

%




\section{CONCLUSION}
In this paper, we explore the use of in-domain depth estimation for end-to-end monocular 3D detection through implicit depth representation learning.
We propose to leverage depth estimation as a proxy task through a multitasking network that encourages representation alignment when either LiDAR data or RGB videos are available in the target domain during training.
Our approach focuses on strengthening representation learning, which is generalizable and complementary to other advances in end-to-end 3D detection algorithms.
\clearpage
\bibliographystyle{IEEEtran}
\bibliography{IEEEabrv,main}

\end{document}